\documentclass[sigconf]{acmart}
\AtBeginDocument{%
  }

\setcopyright{acmlicensed}
\copyrightyear{2026}
\acmYear{2026}
\acmDOI{}
\acmConference[BiAlign '26]{BiAlign Workshop 2026 CHI}{April 15, 2026}{Barcelona, Spain}
\acmBooktitle{Proceedings of the BiAlign Workshop 2026 CHI, April 15, 2026, Barcelona, Spain}
\acmISBN{}

\begin{document}

\title{Multi-System Cognitive Steering for Generative Communication}

\author{Eric Lacosse}
\authornote{shared contribution}
\affiliation{%
  \institution{Champalimaud Centre for Restorative Neurotechnology}
  \institution{Champalimaud Research}
  \city{Lisbon}
  \country{Portugal}}
\email{eric.lacosse@research.fchampalimaud.org}

\author{Pablo Tano}
\authornotemark[1]
\affiliation{%
  \institution{Champalimaud Research}
  \city{Lisbon}
  \country{Portugal}}
\email{pablo.tano@research.fchampalimaud.org}

\author{Daniel C. McNamee}
\affiliation{%
\institution{Champalimaud Research}
  \institution{Champalimaud Centre for Restorative Neurotechnology}
  \city{Lisbon}
  \country{Portugal}}
\email{daniel.mcnamee@}
\email{research.fchampalimaud.org}

\renewcommand{\shortauthors}{Lacosse, Tano et al.}

\begin{abstract}
A fundamental bottleneck for translating thoughts to speech is the high physical cost of motor input forcing a strict tradeoff between communication speed and expressivity.
Generative AI offers a promising solution by expanding highly abbreviated text input into fluent sentences, significantly reducing the motor burden on the user.
However, sparse inputs are inherently ambiguous;
without sufficient context, AI models frequently misinterpret these abbreviations and generate outputs that fail to align with the user's true cognitive intent.
We propose \textit{Cognitive Steering}, a framework that can be leveraged to reconstruct user intent by grounding Large Language Models (LLMs) in multimodal context and applying AI control methods.
We demonstrate that aligning frontier models with selected environmental referents, emotional states, and communication motives on simulated abbreviation benchmarks achieves up to $\sim$88\% top-5 expansion accuracy, doubling no-context baselines.
We also reveal that different types of context require distinct AI control techniques: while physical objects in a scene are best provided through explicit text prompts, abstract factors like emotion may be more effectively managed by directly altering the AI's internal activations.
These findings suggest a hybrid architecture for future interfaces that leverage Cognitive Steering to help users communicate both quickly and accurately, recruiting multiple AI control methods simultaneously to perfectly capture their intended communication goals.
\end{abstract}





\maketitle

\section{Introduction}
Augmentative and Alternative Communication (AAC) systems for users with severe speech and motor impairment, such as those with Amyotrophic Lateral Sclerosis (ALS), suffer from a fundamental bottleneck: they require limited high-cost motor input to express high-bandwidth cognitive intent.
Although humans can speak $\sim$150 words per minute \cite{anumanchipalliSpeechSynthesisNeural2019}, eye-gaze typing interfaces restrict transmission to roughly ten words per minute \cite{majarantaTwentyYearsEye2002}.

\begin{figure*}[!tb]
  \centering
  \includegraphics[width=\linewidth]{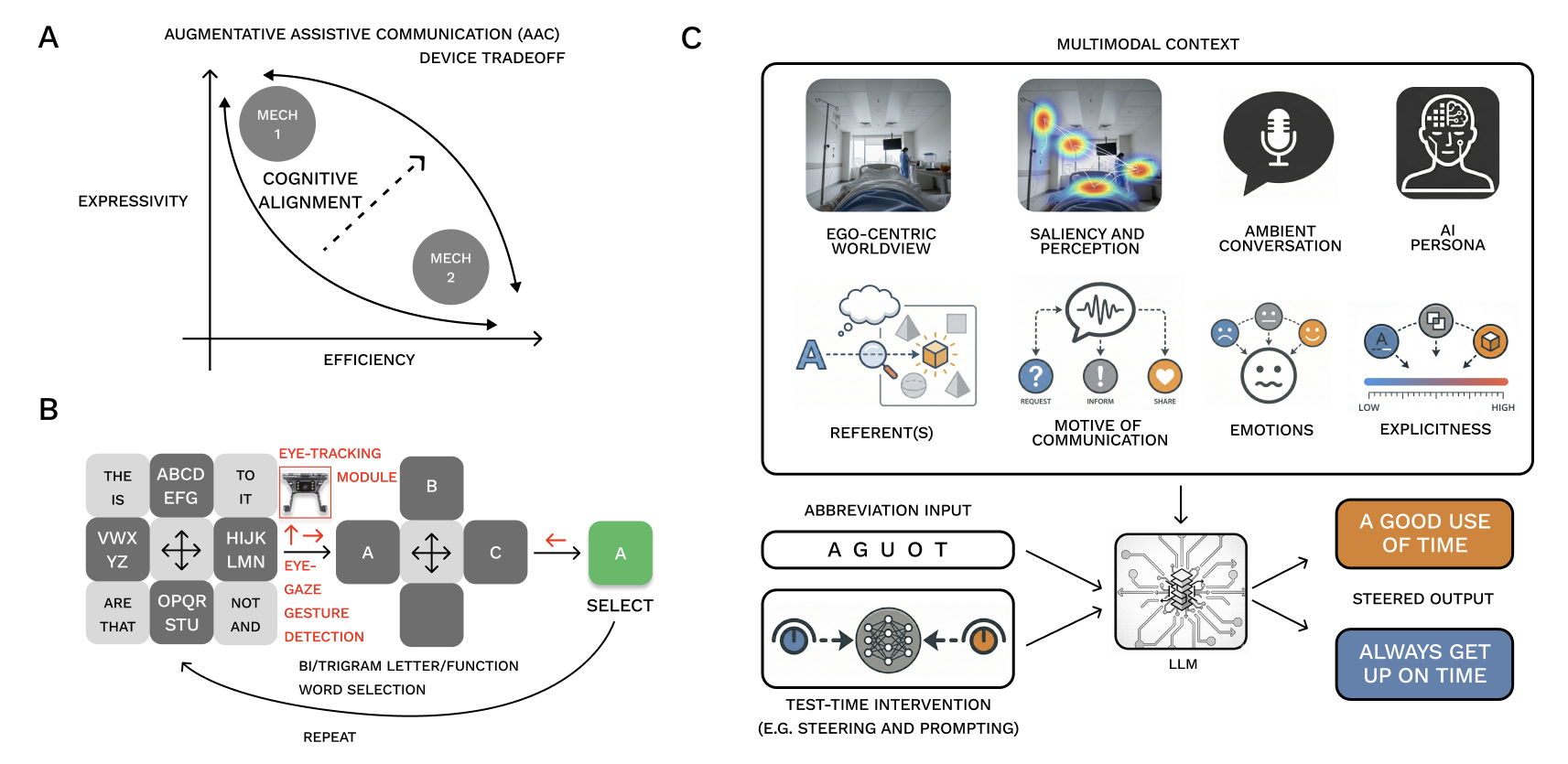}
  \caption{
    \textbf{(A)} The AAC Pareto Frontier.
    Traditional AAC devices force a tradeoff \cite{hoagTradeoffsInformativenessSpeed2004} between expressivity (Communication Expressivity, y-axis)
    and efficiency (Communication Efficiency, x-axis). An example of Mechanism 1 is letter-by-letter typing, which captures nuance but is slow; and an example of Mechanism 2 is preset phrases, which are fast but potentially generic.
    We propose that Cognitive Alignment (CA)---achieved by grounding LLM generation on human intentionality---allows the frontier to be broken as illustrated. By successfully expanding compressed inputs (abbreviations) into the intended utterances, users will better achieve high expressivity without sacrificing efficiency.
    \textbf{(B)} Keyboard expansion and typing based on eye-gaze control.
    A combination of bi-gram and tri-gram allows the selection of letters via a keyboard expansion through directional gestures performed by eye movement.
    Additional high-frequency common words can be placed for easy access on the keyboard for positional gaze detection.
    \textbf{(C)} Contextual Steering. The CCI achieves CA by injecting multimodal context (visual ego-centric view, saliency, audio) and pragmatic elements (referents, motive of communication, emotion) into the LLM inference process. A test-time intervention mechanism (e.g. prompting or steering) uses these signals to steer the LLM, ensuring the abbreviation ``A G U O T'' is expanded into the contextually relevant utterances that steered closer to the intended phrase like (``A GOOD USE OF TIME") or (``ALWAYS GET UP ON TIME").
  }
  \label{fig:1}
\end{figure*}

Generative AI offers a transformative opportunity to bridge this gap by expanding low-bandwidth inputs into fluent natural language (Fig. \ref{fig:1}A) \cite{hoagTradeoffsInformativenessSpeed2004}.
Recent systems allow users to transmit highly compressed signals---extreme abbreviations like ``h-w-t-g-o-a-p'' which are expanded into full sentences (``he wanted to go out and play'') \cite{caiUsingLargeLanguage2024}.
We view the user's abbreviated input not as incomplete data, but as a compressed signal---a ``blurred pattern''---that the AI must use to reconstruct the user's intent \cite{lupyanUnreasonableEffectivenessPattern2026}.
Large Language Models (LLMs), much like human cognition, possess a capacity to reconstruct user intent from such minimal inputs based on contextual cues and grounding \cite{nourChartingTrajectoriesHuman2025}.
In human communication, this reconstructive capacity is largely governed by pragmatics, i.e., the ability to derive intended meaning beyond literal syntax and semantics by relying on shared context, environmental constraints, and conversational maxims \cite{grice1975logic, sperber1986relevance}.

However, relying on LLMs introduces a critical Cognitive Alignment (CA) challenge: without sufficient context, models frequently misinterpret ambiguous user intent \cite{subramonyamBridgingGulfEnvisioning2024}.
For example, a user entering an abbreviation like ``h-w-t-g-o-a-p'' may intend ``he wanted to go out and play'', yet in a different context, the same compressed input could be expanded into the unintended but plausible alternative, e.g., ``how we think governs our actual perception''.
Hence, while abbreviation expansion significantly reduces motor load, it introduces a problem: \emph{Cognitive Misalignment} (CM).
We functionally define CM as the difference between the user's intended message and the AI's interpretation of the user's intent.
This may happen for many reasons, e.g., lack of theory of mind or misinterpreted user intent \cite{subramonyamBridgingGulfEnvisioning2024}, leading the AI to unexpected behavior.

This misalignment between human and AI may be best addressed by treating the LLM not as a digital replica of human cognition, but as a distinct form of intelligence \cite{brattonbenjaminAlignmentAntikythera, haylesBacteriaAIHuman2025}.
While cognitive science has extensively mapped the pragmatic cues humans rely on to resolve ambiguity \cite{clarkUsingLanguage1996}, we lack a comparable understanding of the requirements that enable LLMs to perform similar feats \cite{yingBenchmarkingHumanLikeIntelligence2025}.
There is no guarantee of an isomorphism between the signals that ground human understanding and those that appropriately influence the model's behavior.
An LLM might successfully resolve pragmatic ambiguities that are completely opaque to, or divorced from, human intuition \cite{sreedharanExplainableHumanAIInteraction2024, zahediMentalModelCentricLandscape2022}.
Conversely, an LLM's output might not reflect information that might seem obvious to a human observer/user.
Therefore, rather than imposing an anthropocentric understanding onto these systems, we must empirically interrogate them with minimal assumptions \cite{sharkeyOpenProblemsMechanistic2025}.

This challenge necessitates the analysis presented in this study.
In order to build a capable and robust Cognitive Communication Interface (CCI) for humans and AIs (Fig. \ref{fig:1}B),
we must identify the signals required to achieve CA so the model's behavior outputs what the user intended without costly interventions.
Here, we analyze how different contextual information sources, particularly those from commercially available eye-tracking interfaces \cite{baumannNeonAccuracyTest2023}, can be leveraged to reconstruct abbreviated user inputs across both synthetically generated and real-world datasets, by exploiting pragmatic elements of the context for CA (Fig. \ref{fig:1}C).
We demonstrate that achieving better CA may require distinct technical interventions at test-time inference. While discrete semantic grounding benefits from explicit prompting, high-level intent, like those shaped through latent factors like emotion, can be effectively managed through targeted model interventions of its internal activations at test-time, i.e., activation steering \cite{turnerSteeringLanguageModels2024}, which has been shown to shift the recruitment of latent model concepts \cite{bigelowBeliefDynamicsReveal2025}.
We believe these results will help inform the future engineering and design decisions needed for next-gen CCIs that are bidirectionally aligned between humans and AIs \cite{shen_position_2025}.


\section{Methods.}

\paragraph{Our Generated Dataset.}
Evaluating whether LLMs can resolve ambiguities as a consequence of Cognitive Misalignment requires the use of benchmarks that systematically vary the amounts of pragmatically relevant contexts under which utterances are produced.
Here, we focus on investigating the impact of four \emph{pragmatic factors} (Fig. \ref{fig:1}C):
\begin{enumerate}
\item \emph{Motive of Communication (MOC)}: why the user is speaking and what functional goal are they aiming to achieve? For example, is the user requesting, sharing, or informing \cite{tomasello2010origins}.
\item  \emph{Explicitness}: how concretely has the user expressed themselves? The gradient of underdetermination theory \cite{recanatiDeixisAnaphora2005} defines three levels, namely \emph{elliptical} (i.e., fragmentary, high-context productions such as single words or gapped constructions), \emph{underspecified} (i.e., syntactically complete sentences using pronouns or deictic expressions such as ``that one'' or ``over there''), and \emph{explicit} (i.e., full noun phrases with disambiguating attributes).
\item \emph{Referents}: what are the concrete entities (objects, people, locations) in the physical world that a speaker is talking about.
\item \emph{Emotions}: what is the speaker's latent affective state (e.g., frustration, sadness, neutrality), which acts as a global modulator on the utterance.
\end{enumerate}
Our data generation pipeline proceeds through four sequential stages designed to construct each evaluation datum with complete and separate control over these variables. First, a \emph{scenario configuration} is sampled from a structured ontology of over 235,000 permutations of environmental, embodied, and social factors. These factors include real-world domain (e.g., medical, office), visual complexity (how cluttered a scene is), attentional states (e.g., joint, disengaged, face-to-face),  ego posture (from where the first-person view is taken from), and interlocutor dynamics (the relation between user and interlocutor, epistemic and social). This scenario configuration constrains what a speaker could plausibly say and why with considerable diversity across the generated set.
Second, \emph{referential grounding} uses a prompted LLM to populate each generated scenario with a target object (the semantic anchor of the utterance) and a controlled set of distractor objects in the scene.
Third, \emph{visual scene generation} translates the seeded scenario into structured prompts for a text-to-image model (\texttt{Gemini nano-banana}), producing egocentric scene images that preserve the camera perspective, object composition, and atmospheric cues implied by the scenario factors.
Finally, utterance generation independently and uniformly samples values for the three other pragmatic factors---MOC, Explicitness, and Emotion---then prompts an LLM to produce a naturalistic spoken utterance that is grounded in the scene and shaped by these sampled factors.
The utterance is then reduced to an abbreviation (as explained in the next paragraph), yielding the compressed input signal that the LLM under evaluation must expand back into the intended sentence.

\paragraph{Image-Chat dataset.}
To complement our generated benchmark with an external dataset acquired from human users, we adapt Image-Chat \cite{shusterImageChatEngagingGrounded2020}, a large-scale dataset of grounded human dialogues about images.
In Image-Chat, crowdworkers were each assigned a personality trait (e.g. adventurous, sentimental) and asked to converse about a photograph from the YFCC100m corpus; alternating between Speaker~A and Speaker~B. We analyze only the first two dialogue interactions. We selected 150 random conversations from the dataset, where both the opening (Speaker~A) and response (Speaker~B) utterances contain two to seven words, filtering out entries containing slang, multi-sentence turns, grammatical errors, and pejoratives. The original dataset does not include any description of the image, so we sent the corresponding YFCC100m image to a multimodal LLM (\texttt{Gemini-3-Flash}) to get a scene-description from a first-person point of view, and a list of all distinct visible objects, people, and notable items.

\paragraph{Abbreviation encoding.}
We simulate AAC-style compressed input by reducing each word in a target utterance to its first letter. High-frequency common words (\emph{a, the, is, to, it, are, that, not, you, and}) are given verbatim in square brackets (e.g.\ \texttt{[the] b [is] r h}). The AAC keyboard can include these common words in the interface (Fig. \ref{fig:1}B).

\paragraph{Context layers.}
To systematically test which context information improves abbreviation expansion, we define a hierarchy of \emph{context layers} that incrementally enrich the prompt.
For the generated dataset, the layers are:
L0 (abbreviation only), L1 (+\,communicative intent and motive of communication), L2 (+\,speaker emotion and explicitness), L3 (+\,target and distractor referents), and L4 (+\,scene description). An additional ablation condition ($L4 \setminus L2$) provides referents and scene description but withholds speaker emotions. For the Image-Chat dataset, the layers are:
L0 (abbreviation only; for Speaker B, the preceding turn is always included), L1 (+\,speaker personality), L2 (+\,interlocutor personality), L3 (+\,objects visible in the image), and L4 (+\,full scene description).
We also included the condition $L4 \setminus L2$.

\paragraph{Frontier model evaluation via structured prompting.}
For each context layer, we construct a structured prompt containing the abbreviation, a token-by-token constraint specification (indicating which positions are fixed words vs.\ first-letter hints), and the context features corresponding to that layer.
The prompt instructs the model (\texttt{Gemini-3-Pro}) to return five ranked candidate expansions. For the Image-Chat dataset, separate system prompts are used for Speaker~A (opening turn, no conversational history) and Speaker~B (responding to the preceding turn, which is included in the prompt). In both datasets, we implement a modular template design for prompts that ensures that the only difference between context conditions is the presence or absence of specific context blocks, keeping all other prompt text identical.

\paragraph{Activation steering}
To compare prompting-based alignment with test-time residual intervention, i.e. activation steering \cite{zouRepresentationEngineeringTopDown2025, turnerSteeringLanguageModels2024, rimskySteeringLlama22024, caoPersonalizedSteeringLarge, liuSteeringMultimodalLarge2025}, we use \texttt{OLMo-2-13B-Instruct} \cite{olmo2OLMo22025}.
For each evaluation case, we compute a \emph{steering vector} by running two forward passes, one with the L0 baseline prompt and one with the context-enriched prompt, and taking the difference of the residual-stream activations at the last token position at the target layer: $\mathbf{v}_{\ell} = \mathbf{h}_{\ell}^{\mathrm{context}}[\text{-}1] - \mathbf{h}_{\ell}^{\mathrm{baseline}}[\text{-}1]$. We use the last token since it aggregates the full contextual representation immediately prior to generation.
During generation, this vector is added to the baseline residual at the target layer(s) with a strength $\alpha$: $\mathbf{h}_\ell' = \mathbf{h}_\ell + \alpha\,\mathbf{v}_\ell$. We evaluate two steering configurations: steering into layers 15, 20, and 25 with strength $\alpha=0.5$; and steering into layers 30, 35, and 38 with strength $\alpha=1.0$. Because OLMo-2 is less reliable than the frontier model at unconstrained generation, all decoding is further guided by a per-case regex constraint \cite{willardEfficientGuidedGeneration2023} that enforces the correct word count, preserves bracketed words verbatim, and requires each unknown word to start with the specified letter.

\paragraph{Evaluation metrics.}
We evaluate the degree to which an LLM's reconstructed output recovers the user's intended utterance from an abbreviation signal, using three complementary metrics, each computed at both Top-1 (single best candidate) and Top-5 (best among five candidates). The Top-5 condition simulates the selection-from-list interaction paradigm common in AAC interfaces, where users choose among system-generated candidates rather than accepting a single prediction \cite{caiUsingLargeLanguage2024}.
\emph{Lexical Accuracy} performs a word-level match;
\emph{BLEU} measures the degree of structural and lexical overlap between the reconstruction and the ground truth, is computed with add-one smoothing over $n$-grams of order 1 through $\min(4, \lvert\text{sentence}\rvert)$, with a brevity penalty \cite{papineniBleuMethodAutomatic2002}.
\emph{LLM Judge} asks the \texttt{OLMo-2-13B-Instruct} instance whether the candidate and reference convey equivalent meaning, sentiment, and intent in the current context (full scene description), approximating a holistic pragmatic acceptability judgment.
We consider LLM-Judged equivalency the most complete measure of accuracy, so we refer to it simply as `accuracy' in the rest of the paper, unless otherwise specified.

\section{Results.}

\begin{figure*}[!tb]
  \centering
  \includegraphics[width=0.89\linewidth]{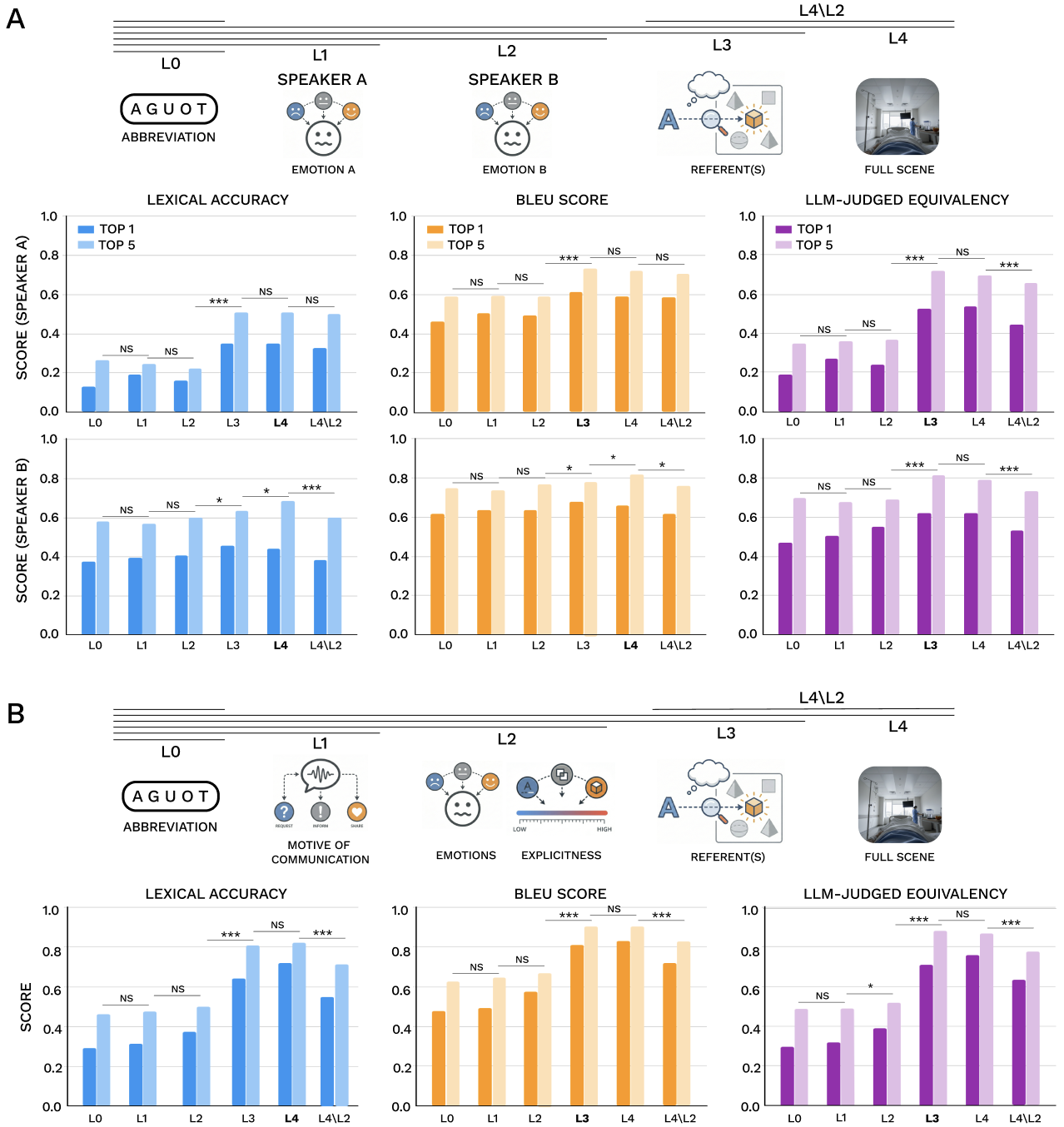}
  \caption{Frontier LLMs achieve up to $\sim 88\%$ top-5 expansion accuracy when aligned to pragmatic context. \textbf{(A)} Image-Chat dataset. (Top) Illustration of the 4 layers. L0 includes no context information and simply asks the LLM to expand the abbreviated sentence. L1 includes the emotion of Speaker A (first dialogue line) and L2 further includes the emotion of Speaker B (second dialogue line). L3 includes referents and L4 the full scene description. (Bottom) Performance for Speaker A (first row) and Speaker B (second row) under different context layers (L0--L4), and in the emotion ablated condition $L4\setminus L2$. The y-axis represents Accuracy, and bars denote Top-1 and Top-5 performance. \textbf{(B)} Generated dataset. (Top) L0 is as in panel A. L1 includes the Motive of Communication (MOC), and L2 includes the speaker emotion and explicitness level. L3 includes target and distractor referents, and L4 the scene description. (Bottom) Performance across context layers and in the emotion ablated condition $L4\setminus L2$. The y-axis represents Accuracy, and bars denote Top-1 and Top-5 performance. Statistical significance was assessed using paired Wilcoxon signed-rank tests (one-sided) with Holm–Bonferroni correction for multiple comparisons. * p < .05, ** p < .01, *** p < .001.}
  \label{fig:results}
  \Description{Panel A shows the performance on the Image-Chat dataset across different context layers (L0 to L4) and an ablated condition, with bar charts for Speaker A and Speaker B. Panel B shows performance on the generated dataset across the same context layers and ablation condition. Bar charts display accuracy metrics with error bars.}
\end{figure*}

\begin{figure*}[!tb]
  \centering
  \includegraphics[width=\linewidth]{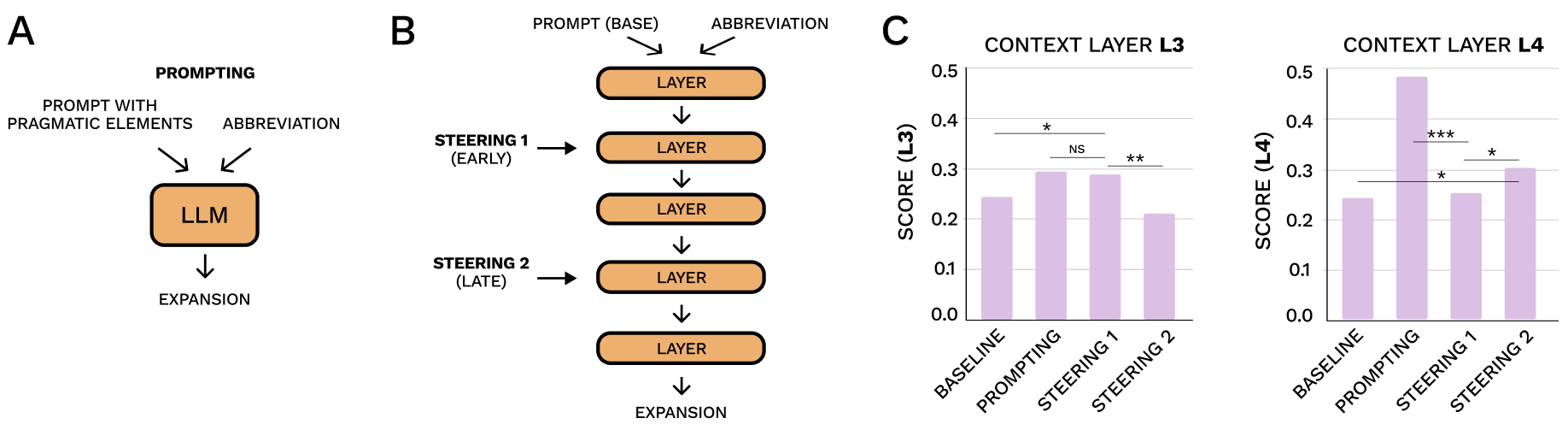}
  \caption{\emph{Cognitive Steering} - Prompting vs.\ test-time steering for alternative context layers. \textbf{(A)} Structured prompting is used to provide pragmatic context features (e.g., MOC, emotions, explicitness, referents) to the model at inference time. \textbf{(B)} Steering procedure: a context-direction vector is added to the residual stream of a chosen layer with varying strength (Steering 1--2), enabling alignment without modifying the prompt. Steering 1 steers layers 15, 20, and 25; Steering 2 steers layers 30, 35, and 38 (\texttt{OLMo-2-13B-Instruct} has 40 layers \cite{olmo2OLMo22025}). \textbf{(C)} Top-5 LLM-judged similarity for alternative steering approaches (bars) on the generated dataset benchmark, for L3 (MOC, emotions, explicitness) and L4 (+Referents) context layer. The y-axis represents Top-5 LLM-judged similarity. Statistical significance was assessed using paired Wilcoxon signed-rank tests (one-sided) with Holm–Bonferroni correction for multiple comparisons. * p < .05, ** p < .01, *** p < .001.}
  \label{fig:steering}
\end{figure*}

\paragraph{Aligned frontier models expand abbreviations with up to 88\% accuracy (Fig.~\ref{fig:results})} Frontier LLMs have an exceptional ability to reconstruct a sentence given a corrupted version \cite{lupyanUnreasonableEffectivenessPattern2026}.
We find that this capability extends to abbreviation expansion: given only the user's abbreviation, frontier models can often infer the intended full phrase, even when the abbreviation occurs in the first or second turn of a dialogue.
This inference is not purely lexical. Many abbreviations are compatible with multiple plausible expansions, and disambiguation requires pragmatic grounding (e.g., who/what is being referred to, the speaker's affect, and the desired communicative intent).
To test whether such grounding is necessary, we provide the model with controlled access to context elements (Fig.~\ref{fig:1}C) via structured prompting (Fig.~\ref{fig:steering}A), and evaluate performance as these elements are included or ablated.
Throughout, we use \texttt{Gemini-3-Pro} as our frontier model, and defer comparisons to alternative alignment mechanisms in open-source models to the next subsection.

In the Image-Chat dataset (Fig.~\ref{fig:results}A), the first dialogue line (Speaker A; first row) is the most underdetermined setting: there is no conversational history, so the model must rely on extra-linguistic cues to resolve ambiguous shorthand.
Exploiting the context, the LLM interpreter achieves 73\% abbreviation-expansion performance (see L3, top-5 LLM-judged equivalency, from now on we refer to this performance metric for the analysis), doubling the performance of the no-context baseline (L0).
A Friedman test confirmed a significant effect of context layer on decoding performance ($\chi^2(4) = 23.98$, $p < .001$).
Only the step L2 to L3 showed significant improvement for top-5 performance, consistent with the hypothesis that abbreviation expansion in situ is often a reference-resolution problem.
However, referents alone are not sufficient for peak performance: ablating the speaker's emotions ($L4\setminus L2$) produces a consistent drop ($\Delta = -.05$, $W = 72{,}385$, $p < .001$, $r = .15$; one-sided Wilcoxon signed-rank), indicating that affective state contributes information that the model uses to select between competing expansions (e.g., neutral vs.\ frustrated or urgent formulations). By contrast, the second dialogue line (Speaker B; second row) is partially constrained by the immediately preceding turn, and therefore admits higher performance even without explicit context features (compare \textit{L0} between Speaker A and B). Nevertheless, adding referents and emotions in the prompt still yields an additional $\sim$13\% top-5 LLM-judged gain over the baseline, and this gain largely disappears when emotions are removed ($L4\setminus L2$, $\Delta = -.09$, $W = 3{,}906$, $p < .001$, $r = .33$; one-sided Wilcoxon signed-rank), suggesting that affect provides complementary signal rather than redundant information.

The same pattern holds in the generated dataset (Fig.~\ref{fig:results}B).
When supplied with pragmatic context, the frontier LLM reconstructs the intended phrase from the abbreviation with $88\%$ top-5 accuracy, even in the absence of any prior conversational content.
Since top-5 corresponds to a realistic AAC interaction loop in which the user selects from candidate expansions, these results imply that context-dependent alignment can materially improve usability:
it increases the chance that the intended message appears among a small set of options, while preserving the efficiency benefits of abbreviated input.
A Friedman test confirmed a significant effect of context layer on decoding performance ($\chi^2(4) = 126.67$, $p < .001$).
Ablating the MOC, emotion and explicitness ($L4\setminus L2$) significantly lowers performance over the full L4 ($\Delta = -.170$, $W = 240.0$, $p < .001$, $r = .65$).

\paragraph{The efficiency of alignment approaches depends on context type (Fig.~\ref{fig:steering})}.
We found that the efficiency of alternative approaches for Cognitive Alignment depends on which dimension of the context is being considered (e.g. emotions vs. referents).
Latent steering \cite{turnerSteeringLanguageModels2024, vuAngularSteeringBehavior2025, nguyenActivationSteeringFeedback2025, liuSteeringMultimodalLarge2025, banayeeanzadePsychologicalSteeringLLMs2025, wuImprovedRepresentationSteering2025, zouRepresentationEngineeringTopApproach2023} is a promising approach for alignment, an alternative to prompting, by which a vector is added to the residual activity of a target layer(s) at test-time, in order to influence the model's behavior.
Evaluating steering in proprietary models is not possible as it requires accessing intermediate activations during inference, so for this analysis we use the open-source \texttt{OLMo-2-13B-Instruct} model \cite{olmo2OLMo22025}, evaluated on the generated dataset.
This model achieves significantly lower performance than the frontier model (compare with Fig.~\ref{fig:results}B), but it allows us to compare test-time intervention approaches for different context types. We compare prompting (Fig.~\ref{fig:steering}A) with steering on early or later layers (Steering 1 and 2 in Fig.~\ref{fig:steering}B).

We found that aligning the L3 context layer in Fig.~\ref{fig:results}B benefits from different approaches than aligning to L4 in Fig.~\ref{fig:results}B (Fig.~\ref{fig:steering}C). Alignment to MOC, Emotions and Explicitness (L3) benefits from steering early layers, which performs similarly to prompting. In contrast, alignment to referents (L4) is best achieved by prompting, or by steering a late layer. Intuitively, MOC, affect, and explicitness largely shape \emph{how} an utterance should be produced. Steering early layers can act like a subtle prior: it nudges the model into the appropriate communicative mode without overriding the semantic or lexical content. In contrast, resolving referents is primarily about \emph{what} the utterance is about: selecting the correct entities and filling in missing arguments from the context. Directly providing referents in the prompt is highly effective because it makes the relevant entities explicit in the model's immediate context, similarly to decisively shifting token-level logits toward the intended entity using late-layer steering.

\section{Discussion.}
We presented a novel multi-system Cognitive Steering framework for generative communication that relies on test-time intervention methods.
We investigated structured context ablations for its use in LLMs.
By treating user input as a compressed signal to be reconstructed, we demonstrate that LLMs have the potential to enable high-bandwidth communication from extremely limited motor inputs (e.g., selecting abbreviations).
However, this capability is contingent upon solving the problem of Cognitive Misalignment whereby the LLM lacks the necessary context to resolve user intention.
Our results suggest that solving misalignment requires grounding the LLM in highly specific pragmatic contexts, e.g., visual attention, emotional state, and communicative intent, that human listeners naturally utilize.
With appropriate context, like that available from commercially wearable eye-trackers---those with world-cameras and microphones to catch and diarize conversations---we demonstrate that frontier LLMs can reconstruct intended utterances from highly compressed abbreviations with up to 88\% top-5 accuracy (Fig.~\ref{fig:results}).
In particular, we found that (visual) referents--information readily available via the same eye-tracking hardware used for typing--are the most significant driver of disambiguation (Fig.\ \ref{fig:results}).
This suggests a promising direction for future work where eye-gaze interfaces do not merely act as cursors for input, but as semantic sensors and selectors, actively logging attended objects to engineer context \cite{bulling2010eye}.

While visual context (referents and scene descriptions) provided the most significant boost in lexical accuracy (Fig. \ref{fig:results}), the small but consistent performance drop observed in the emotion-ablated condition ($L4 \setminus L2$) suggests that semantic grounding (``what acts upon what'') can be paired with pragmatic grounding (``how and why the speaker feels'') to further narrow the search space of plausible expansions.
This aligns with theories of human pragmatics suggesting that intent recognition is often distinct from reference resolution \cite{tomasello2010origins}.
For the user, this suggests that sensing the user's latent mental state may further improve expansion accuracy.

Our comparison of structured prompting versus activation steering (Fig. \ref{fig:steering}) suggests that different dimensions of communication map to different intervention points in the model architecture. Specific objects in a scene benefit from explicit prompting or late-layer steering.
Conversely, abstract features like MOC and Emotion were effectively managed via early-layer steering.
While observed in a specific open-source model, this supports the view that early layers process high-level style and global constraints, while later layers resolve specific token identities, though this needs verification in more capable frontier models before generalizing to all pretrained transformers.
This dissociation suggests a path toward hybrid CCIs:
using discrete prompts to ground the model in the physical world (via object detection) while using steering vectors to modulate the ``voice'' and intent of the output.
Furthermore, treating in-context learning (prompting) as evidence accumulation and steering as a shift in prior beliefs provides a unified Bayesian framework for precisely predicting how these combined interventions will influence the model's final generated output \cite{bigelowBeliefDynamicsReveal2025}.
Such an architecture is reflected in dual systems theory of human brain function whereby the fast, implicit System 1 and flexible, explicit System 2 divide the algorithmic labor of controlling thought and behavior \cite{dolanGoalsHabitsBrain2013}.
A deeper understanding of this analogy may open further opportunities for bidirectional alignment between AIs and humans.

\section{Limitations and Future Work.}
(1) Due to the closed nature of frontier models (\texttt{Gemini-3-Pro}), our activation steering experiments were restricted to open-source models (\texttt{OLMo-2-13B-Instruct} \cite{olmo2OLMo22025}), which show significantly lower baseline performance. (2)
While we investigated controlled ablation of context factors, our evaluation relies on synthetic data and conversations from crowdworkers, which may not accurately reflect the linguistic patterns, errors, or true cognitive intent of individuals, especially those with severe motor and speech impairments using AAC devices.
We need to deploy a real system under realistic conditions and to investigate whether simulated model performance reflects this. (3)
Our simple first-letter abbreviation method assumes perfect spelling and consistent encoding, which may not hold true for real-world AAC users who often employ idiosyncratic abbreviations, make typographical errors, or may introduce a significant cognitive load. (4)
While frontier model results appear very promising, achieving the latency required for fluent conversation remains an engineering challenge given a gesture-based interface, e.g., Fig. \ref{fig:1}B. (5)
While our results show a benefit to including emotional context, reliably capturing a user's latent mental state in real-time AAC systems remains a significant technical and practical challenge. (6)
We did not include n-gram baselines in our comparisons, which could further contextualize the performance gains of our LLM-based approach. (7)
Finally, the introduction of AI that infers intent raises ethical questions regarding agency \cite{jordanCollectivistEconomicPerspective2025, instituteNobleUsesAI2024}.
Preserving this agency is paramount: future systems must be designed to speak \textit{for} the user, never \textit{over} them \cite{thacker2020age}.


\section{Conclusion.}
Generative AI transforms AAC into a cognitive bidirectional alignment task \cite{shen_position_2025}, involving a tradeoff between communication efficiency and expressivity exists.
We conclude that by coupling low-bandwidth input with rich specific context and through applying the appropriate AI control (e.g., steering mechanisms), we can construct interfaces deeply attuned to the cognitive reality of the user thus overcoming Cognitive Misalignment and transform minimal input into maximum expression.

\section{Acknowledgments}
We would especially like to thank Elizaveta Gurina for her feedback on this manuscript.
E.L., P.T., D.M. thank the Champalimaud Foundation and the Portuguese Recovery and Resilience Plan (project number 62) for funding.
Google Research Credits 423033108.

\bibliographystyle{ACM-Reference-Format}
\bibliography{sample-base}

@incollection{grice1975logic,
  title={Logic and conversation},
  author={Grice, Herbert P},
  booktitle={Speech acts},
  pages={41--58},
  year={1975},
  publisher={Brill}
}

@book{sperber1986relevance,
  title={Relevance: Communication and cognition},
  author={Sperber, Dan and Wilson, Deirdre},
  volume={142},
  year={1986},
  publisher={Harvard University Press Cambridge, MA}
}

@article{anumanchipalliSpeechSynthesisNeural2019,
  title = {Speech Synthesis from Neural Decoding of Spoken Sentences},
  author = {Anumanchipalli, Gopala K. and Chartier, Josh and Chang, Edward F.},
  year = 2019,
  month = apr,
  journal = {Nature},
  volume = {568},
  number = {7753},
  pages = {493--498},
  issn = {1476-4687},
  doi = {10.1038/s41586-019-1119-1},
  langid = {english},
  pmcid = {PMC9714519},
  pmid = {31019317}
}

@misc{banayeeanzadePsychologicalSteeringLLMs2025,
  title = {Psychological {{Steering}} in {{LLMs}}: {{An Evaluation}} of {{Effectiveness}} and {{Trustworthiness}}},
  shorttitle = {Psychological {{Steering}} in {{LLMs}}},
  author = {Banayeeanzade, Amin and Tak, Ala N. and Bahrani, Fatemeh and Bolourani, Anahita and Blas, Leonardo and Ferrara, Emilio and Gratch, Jonathan and Karimireddy, Sai Praneeth},
  year = 2025,
  month = oct,
  number = {arXiv:2510.04484},
  eprint = {2510.04484},
  primaryclass = {cs},
  publisher = {arXiv},
  doi = {10.48550/arXiv.2510.04484},
  urldate = {2025-10-19},
  archiveprefix = {arXiv}
}

@techreport{baumannNeonAccuracyTest2023,
  title = {Neon {{Accuracy Test Report}}},
  author = {Baumann, Chris and Dierkes, Kai},
  year = 2023,
  month = dec,
  institution = {Pupil Labs},
  doi = {10.5281/ZENODO.10420388},
  urldate = {2026-02-12},
  langid = {english}
}

@misc{bigelowBeliefDynamicsReveal2025,
  title = {Belief {{Dynamics Reveal}} the {{Dual Nature}} of {{In-Context Learning}} and {{Activation Steering}}},
  author = {Bigelow, Eric and Wurgaft, Daniel and Wang, YingQiao and Goodman, Noah and Ullman, Tomer and Tanaka, Hidenori and Lubana, Ekdeep Singh},
  year = 2025,
  month = nov,
  number = {arXiv:2511.00617},
  eprint = {2511.00617},
  primaryclass = {cs},
  publisher = {arXiv},
  doi = {10.48550/arXiv.2511.00617},
  urldate = {2025-11-15},
  archiveprefix = {arXiv}
}

@misc{brattonbenjaminAlignmentAntikythera,
  title = {After {{Alignment}} \textbar{} {{Antikythera}}},
  author = {Bratton, Benjamin},
  urldate = {2025-08-21},
  howpublished = {https://afteralignment.antikythera.org/},
  langid = {english}
}

@article{caoPersonalizedSteeringLarge,
  title = {Personalized {{Steering}} of {{Large Language Models}}: {{Versatile Steering Vectors Through Bi-directional Preference Optimization}}},
  author = {Cao, Yuanpu and Zhang, Tianrong and Cao, Bochuan and Yin, Ziyi and Lin, Lu and Ma, Fenglong and Chen, Jinghui},
  langid = {english}
}

@article{hoagTradeoffsInformativenessSpeed2004,
  title = {Trade-Offs between Informativeness and Speed of Message Delivery in Augmentative and Alternative Communication},
  author = {Hoag, Linda A. and Bedrosian, Jan L. and McCoy, Kathleen F. and Johnson, Dallas E.},
  year = 2004,
  month = dec,
  journal = {Journal of speech, language, and hearing research: JSLHR},
  volume = {47},
  number = {6},
  pages = {1270--1285},
  issn = {1092-4388},
  doi = {10.1044/1092-4388(2004/096)},
  langid = {english},
  pmid = {15842010}
}

@misc{instituteNobleUsesAI2024,
  title = {On the {{Noble Uses}} of {{AI}}},
  author = {Institute, Cosmos},
  year = 2024,
  month = oct,
  urldate = {2026-02-15},
  howpublished = {https://blog.cosmos-institute.org/p/on-the-noble-uses-of-ai?utm\_medium=ios\&triedRedirect=true},
  langid = {english}
}

@misc{liuSteeringMultimodalLarge2025,
  title = {Steering {{Multimodal Large Language Models Decoding}} for {{Context-Aware Safety}}},
  author = {Liu, Zheyuan and Xu, Zhangchen and Dou, Guangyao and Yuan, Xiangchi and Tan, Zhaoxuan and Poovendran, Radha and Jiang, Meng},
  year = 2025,
  month = sep,
  number = {arXiv:2509.19212},
  eprint = {2509.19212},
  primaryclass = {cs},
  publisher = {arXiv},
  doi = {10.48550/arXiv.2509.19212},
  urldate = {2025-11-21},
  archiveprefix = {arXiv}
}

@inproceedings{majarantaTwentyYearsEye2002,
  title = {Twenty Years of Eye Typing: Systems and Design Issues},
  shorttitle = {Twenty Years of Eye Typing},
  booktitle = {Proceedings of the 2002 Symposium on {{Eye}} Tracking Research \& Applications},
  author = {Majaranta, P{\"a}ivi and R{\"a}ih{\"a}, Kari-Jouko},
  year = 2002,
  month = mar,
  series = {{{ETRA}} '02},
  pages = {15--22},
  publisher = {Association for Computing Machinery},
  address = {New York, NY, USA},
  doi = {10.1145/507072.507076},
  urldate = {2026-02-12},
  isbn = {978-1-58113-467-4}
}

@misc{nourChartingTrajectoriesHuman2025,
  title = {Charting Trajectories of Human Thought Using Large Language Models},
  author = {Nour, Matthew M. and McNamee, Daniel C. and Fradkin, Isaac and Dolan, Raymond J.},
  year = 2025,
  month = sep,
  number = {arXiv:2509.14455},
  eprint = {2509.14455},
  primaryclass = {q-bio},
  publisher = {arXiv},
  doi = {10.48550/arXiv.2509.14455},
  urldate = {2026-02-13},
  archiveprefix = {arXiv}
}

@misc{olmo2OLMo22025,
  title = {2 {{OLMo}} 2 {{Furious}}},
  author = {OLMo, Team and Walsh, Pete and Soldaini, Luca and Groeneveld, Dirk and Lo, Kyle and Arora, Shane and Bhagia, Akshita and Gu, Yuling and Huang, Shengyi and Jordan, Matt and Lambert, Nathan and Schwenk, Dustin and Tafjord, Oyvind and Anderson, Taira and Atkinson, David and Brahman, Faeze and Clark, Christopher and Dasigi, Pradeep and Dziri, Nouha and Ettinger, Allyson and Guerquin, Michal and Heineman, David and Ivison, Hamish and Koh, Pang Wei and Liu, Jiacheng and Malik, Saumya and Merrill, William and Miranda, Lester James V. and Morrison, Jacob and Murray, Tyler and Nam, Crystal and Poznanski, Jake and Pyatkin, Valentina and Rangapur, Aman and Schmitz, Michael and Skjonsberg, Sam and Wadden, David and Wilhelm, Christopher and Wilson, Michael and Zettlemoyer, Luke and Farhadi, Ali and Smith, Noah A. and Hajishirzi, Hannaneh},
  year = 2025,
  month = oct,
  number = {arXiv:2501.00656},
  eprint = {2501.00656},
  primaryclass = {cs},
  publisher = {arXiv},
  doi = {10.48550/arXiv.2501.00656},
  urldate = {2026-02-12},
  archiveprefix = {arXiv}
}

@inproceedings{rimskySteeringLlama22024,
  title = {Steering {{Llama}} 2 via {{Contrastive Activation Addition}}},
  booktitle = {Proceedings of the 62nd {{Annual Meeting}} of the {{Association}} for {{Computational Linguistics}} ({{Volume}} 1: {{Long Papers}})},
  author = {Rimsky, Nina and Gabrieli, Nick and Schulz, Julian and Tong, Meg and Hubinger, Evan and Turner, Alexander},
  year = 2024,
  pages = {15504--15522},
  publisher = {Association for Computational Linguistics},
  address = {Bangkok, Thailand},
  doi = {10.18653/v1/2024.acl-long.828},
  urldate = {2025-08-10},
  langid = {english}
}

@misc{sharkeyOpenProblemsMechanistic2025,
  title = {Open {{Problems}} in {{Mechanistic Interpretability}}},
  author = {Sharkey, Lee and Chughtai, Bilal and Batson, Joshua and Lindsey, Jack and Wu, Jeff and Bushnaq, Lucius and {Goldowsky-Dill}, Nicholas and Heimersheim, Stefan and Ortega, Alejandro and Bloom, Joseph and Biderman, Stella and {Garriga-Alonso}, Adria and Conmy, Arthur and Nanda, Neel and Rumbelow, Jessica and Wattenberg, Martin and Schoots, Nandi and Miller, Joseph and Michaud, Eric J. and Casper, Stephen and Tegmark, Max and Saunders, William and Bau, David and Todd, Eric and Geiger, Atticus and Geva, Mor and Hoogland, Jesse and Murfet, Daniel and McGrath, Tom},
  year = 2025,
  month = jan,
  number = {arXiv:2501.16496},
  eprint = {2501.16496},
  primaryclass = {cs},
  publisher = {arXiv},
  doi = {10.48550/arXiv.2501.16496},
  urldate = {2025-04-20},
  archiveprefix = {arXiv},
  langid = {english}
}

@inproceedings{shusterImageChatEngagingGrounded2020,
  title = {Image-{{Chat}}: {{Engaging Grounded Conversations}}},
  shorttitle = {Image-{{Chat}}},
  booktitle = {Proceedings of the 58th {{Annual Meeting}} of the {{Association}} for {{Computational Linguistics}}},
  author = {Shuster, Kurt and Humeau, Samuel and Bordes, Antoine and Weston, Jason},
  editor = {Jurafsky, Dan and Chai, Joyce and Schluter, Natalie and Tetreault, Joel},
  year = 2020,
  month = jul,
  pages = {2414--2429},
  publisher = {Association for Computational Linguistics},
  address = {Online},
  doi = {10.18653/v1/2020.acl-main.219},
  urldate = {2026-02-10}
}

@misc{turnerSteeringLanguageModels2024,
  title = {Steering {{Language Models With Activation Engineering}}},
  author = {Turner, Alexander Matt and Thiergart, Lisa and Leech, Gavin and Udell, David and Vazquez, Juan J. and Mini, Ulisse and MacDiarmid, Monte},
  year = 2024,
  month = oct,
  number = {arXiv:2308.10248},
  eprint = {2308.10248},
  primaryclass = {cs},
  publisher = {arXiv},
  doi = {10.48550/arXiv.2308.10248},
  urldate = {2025-08-01},
  archiveprefix = {arXiv}
}

@misc{vuAngularSteeringBehavior2025,
  title = {Angular {{Steering}}: {{Behavior Control}} via {{Rotation}} in {{Activation Space}}},
  shorttitle = {Angular {{Steering}}},
  author = {Vu, Hieu M. and Nguyen, Tan M.},
  year = 2025,
  month = oct,
  number = {arXiv:2510.26243},
  eprint = {2510.26243},
  primaryclass = {cs},
  publisher = {arXiv},
  doi = {10.48550/arXiv.2510.26243},
  urldate = {2025-11-21},
  archiveprefix = {arXiv}
}

@misc{willardEfficientGuidedGeneration2023,
  title = {Efficient {{Guided Generation}} for {{Large Language Models}}},
  author = {Willard, Brandon T. and Louf, R{\'e}mi},
  year = 2023,
  month = aug,
  number = {arXiv:2307.09702},
  eprint = {2307.09702},
  primaryclass = {cs},
  publisher = {arXiv},
  doi = {10.48550/arXiv.2307.09702},
  urldate = {2026-03-19},
  archiveprefix = {arXiv}
}

@misc{wuImprovedRepresentationSteering2025,
  title = {Improved {{Representation Steering}} for {{Language Models}}},
  author = {Wu, Zhengxuan and Yu, Qinan and Arora, Aryaman and Manning, Christopher D. and Potts, Christopher},
  year = 2025,
  month = may,
  number = {arXiv:2505.20809},
  eprint = {2505.20809},
  primaryclass = {cs},
  publisher = {arXiv},
  doi = {10.48550/arXiv.2505.20809},
  urldate = {2025-08-01},
  archiveprefix = {arXiv}
}

@misc{nguyenActivationSteeringFeedback2025,
	title = {Activation {Steering} with a {Feedback} {Controller}},
	url = {http://arxiv.org/abs/2510.04309},
	doi = {10.48550/arXiv.2510.04309},
	urldate = {2025-10-27},
	publisher = {arXiv},
	author = {Nguyen, Dung V. and Vu, Hieu M. and Pham, Nhi Y. and Zhang, Lei and Nguyen, Tan M.},
	month = oct,
	year = {2025},
	note = {arXiv:2510.04309 [cs]},
}

@misc{zouRepresentationEngineeringTopDown2025,
	title = {Representation {Engineering}: {A} {Top}-{Down} {Approach} to {AI} {Transparency}},
	shorttitle = {Representation {Engineering}},
	url = {http://arxiv.org/abs/2310.01405},
	doi = {10.48550/arXiv.2310.01405},
	urldate = {2025-04-20},
	publisher = {arXiv},
	author = {Zou, Andy and Phan, Long and Chen, Sarah and Campbell, James and Guo, Phillip and Ren, Richard and Pan, Alexander and Yin, Xuwang and Mazeika, Mantas and Dombrowski, Ann-Kathrin and Goel, Shashwat and Li, Nathaniel and Byun, Michael J. and Wang, Zifan and Mallen, Alex and Basart, Steven and Koyejo, Sanmi and Song, Dawn and Fredrikson, Matt and Kolter, J. Zico and Hendrycks, Dan},
	month = mar,
	year = {2025},
	note = {arXiv:2310.01405 [cs]},
}

@book{thacker2020age,
  title={The age of AI: Artificial intelligence and the future of humanity},
  author={Thacker, Jason},
  year={2020},
  publisher={HarperChristian+ ORM}
}

@article{dolanGoalsHabitsBrain2013,
	title = {Goals and {Habits} in the {Brain}},
	volume = {80},
	issn = {0896-6273},
	url = {https://pmc.ncbi.nlm.nih.gov/articles/PMC3807793/},
	doi = {10.1016/j.neuron.2013.09.007},
	number = {2},
	urldate = {2026-02-13},
	journal = {Neuron},
	author = {Dolan, Ray J. and Dayan, Peter},
	month = oct,
	year = {2013},
	pages = {312--325},
}

@incollection{recanatiDeixisAnaphora2005,
	title = {Deixis and {Anaphora}},
	isbn = {978-0-19-925152-0},
	url = {https://doi.org/10.1093/acprof:oso/9780199251520.003.0007},
	doi = {10.1093/acprof:oso/9780199251520.003.0007},
	urldate = {2026-02-13},
	booktitle = {Semantics versus {Pragmatics}},
	publisher = {Oxford University Press},
	author = {Recanati, François},
	editor = {Szabo, Zoltan Gendler},
	month = jan,
	year = {2005},
	pages = {0},
}

@book{tomasello2010origins,
  title={Origins of human communication},
  author={Tomasello, Michael},
  year={2010},
  publisher={MIT press}
}

@article{bulling2010eye,
  title={Eye movement analysis for activity recognition using electrooculography},
  author={Bulling, Andreas and Ward, Jamie A and Gellersen, Hans and Tr{\"o}ster, Gerhard},
  journal={IEEE transactions on pattern analysis and machine intelligence},
  volume={33},
  number={4},
  pages={741--753},
  year={2010},
  publisher={IEEE}
}

@misc{zouRepresentationEngineeringTopApproach2023,
	title = {Representation {Engineering}: {A} {Top}-{Down} {Approach} to {AI} {Transparency}},
	shorttitle = {Representation {Engineering}},
	url = {http://arxiv.org/abs/2310.01405},
	urldate = {2024-11-03},
	publisher = {arXiv},
	author = {Zou, Andy and Phan, Long and Chen, Sarah and Campbell, James and Guo, Phillip and Ren, Richard and Pan, Alexander and Yin, Xuwang and Mazeika, Mantas and Dombrowski, Ann-Kathrin and Goel, Shashwat and Li, Nathaniel and Byun, Michael J. and Wang, Zifan and Mallen, Alex and Basart, Steven and Koyejo, Sanmi and Song, Dawn and Fredrikson, Matt and Kolter, J. Zico and Hendrycks, Dan},
	month = oct,
	year = {2023},
	note = {arXiv:2310.01405},
}

@misc{subramonyamBridgingGulfEnvisioning2024,
	title = {Bridging the {Gulf} of {Envisioning}: {Cognitive} {Design} {Challenges} in {LLM} {Interfaces}},
	shorttitle = {Bridging the {Gulf} of {Envisioning}},
	url = {http://arxiv.org/abs/2309.14459},
	doi = {10.48550/arXiv.2309.14459},
	urldate = {2025-01-07},
	publisher = {arXiv},
	author = {Subramonyam, Hariharan and Pea, Roy and Pondoc, Christopher Lawrence and Agrawala, Maneesh and Seifert, Colleen},
	month = mar,
	year = {2024},
	note = {arXiv:2309.14459 [cs]},
}

@article{caiUsingLargeLanguage2024,
	title = {Using large language models to accelerate communication for eye gaze typing users with {ALS}},
	volume = {15},
	copyright = {2024 The Author(s)},
	issn = {2041-1723},
	url = {https://www.nature.com/articles/s41467-024-53873-3},
	doi = {10.1038/s41467-024-53873-3},
	language = {en},
	number = {1},
	urldate = {2025-03-01},
	journal = {Nature Communications},
	publisher = {Nature Publishing Group},
	author = {Cai, Shanqing and Venugopalan, Subhashini and Seaver, Katie and Xiao, Xiang and Tomanek, Katrin and Jalasutram, Sri and Morris, Meredith Ringel and Kane, Shaun and Narayanan, Ajit and MacDonald, Robert L. and Kornman, Emily and Vance, Daniel and Casey, Blair and Gleason, Steve M. and Nelson, Philip Q. and Brenner, Michael P.},
	month = nov,
	year = {2024},
	pages = {9449},
}

@misc{yingBenchmarkingHumanLikeIntelligence2025,
	title = {On {Benchmarking} {Human}-{Like} {Intelligence} in {Machines}},
	url = {http://arxiv.org/abs/2502.20502},
	doi = {10.48550/arXiv.2502.20502},
	urldate = {2025-04-01},
	publisher = {arXiv},
	author = {Ying, Lance and Collins, Katherine M. and Wong, Lionel and Sucholutsky, Ilia and Liu, Ryan and Weller, Adrian and Shu, Tianmin and Griffiths, Thomas L. and Tenenbaum, Joshua B.},
	month = feb,
	year = {2025},
	note = {arXiv:2502.20502 [cs]},
}

@misc{jordanCollectivistEconomicPerspective2025,
	title = {A {Collectivist}, {Economic} {Perspective} on {AI}},
	url = {http://arxiv.org/abs/2507.06268},
	doi = {10.48550/arXiv.2507.06268},
	urldate = {2025-08-01},
	publisher = {arXiv},
	author = {Jordan, Michael I.},
	month = jul,
	year = {2025},
	note = {arXiv:2507.06268 [cs]},
}

@inproceedings{papineniBleuMethodAutomatic2002,
	address = {Philadelphia, Pennsylvania, USA},
	title = {Bleu: a {Method} for {Automatic} {Evaluation} of {Machine} {Translation}},
	shorttitle = {Bleu},
	url = {https://aclanthology.org/P02-1040/},
	doi = {10.3115/1073083.1073135},
	urldate = {2025-09-18},
	booktitle = {Proceedings of the 40th {Annual} {Meeting} of the {Association} for {Computational} {Linguistics}},
	publisher = {Association for Computational Linguistics},
	author = {Papineni, Kishore and Roukos, Salim and Ward, Todd and Zhu, Wei-Jing},
	editor = {Isabelle, Pierre and Charniak, Eugene and Lin, Dekang},
	month = jul,
	year = {2002},
	pages = {311--318},
}

@misc{zahediMentalModelCentricLandscape2022,
	title = {A {Mental}-{Model} {Centric} {Landscape} of {Human}-{AI} {Symbiosis}},
	url = {http://arxiv.org/abs/2202.09447},
	doi = {10.48550/arXiv.2202.09447},
	urldate = {2025-09-28},
	publisher = {arXiv},
	author = {Zahedi, Zahra and Sreedharan, Sarath and Kambhampati, Subbarao},
	month = feb,
	year = {2022},
	note = {arXiv:2202.09447 [cs]},
}

@misc{sreedharanExplainableHumanAIInteraction2024,
	title = {Explainable {Human}-{AI} {Interaction}: {A} {Planning} {Perspective}},
	shorttitle = {Explainable {Human}-{AI} {Interaction}},
	url = {http://arxiv.org/abs/2405.15804},
	doi = {10.48550/arXiv.2405.15804},
	urldate = {2025-09-29},
	publisher = {arXiv},
	author = {Sreedharan, Sarath and Kulkarni, Anagha and Kambhampati, Subbarao},
	month = may,
	year = {2024},
	note = {arXiv:2405.15804 [cs]},
}

@misc{lupyanUnreasonableEffectivenessPattern2026,
	title = {The unreasonable effectiveness of pattern matching},
	url = {http://arxiv.org/abs/2601.11432},
	doi = {10.48550/arXiv.2601.11432},
	urldate = {2026-01-24},
	publisher = {arXiv},
	author = {Lupyan, Gary and Arcas, Blaise Agüera y},
	month = jan,
	year = {2026},
	note = {arXiv:2601.11432 [cs]},
}

@book{haylesBacteriaAIHuman2025,
	title = {Bacteria to {AI}: {Human} {Futures} with {Our} {Nonhuman} {Symbionts}},
	isbn = {978-0-226-83746-8},
	shorttitle = {Bacteria to {AI}},
	language = {en},
	publisher = {University of Chicago Press},
	author = {Hayles, N. Katherine},
	month = feb,
	year = {2025},
	note = {Google-Books-ID: hZEVEQAAQBAJ},
}

@book{clarkUsingLanguage1996,
	title = {Using {Language}},
	isbn = {978-0-521-56745-9},
	language = {en},
	publisher = {Cambridge University Press},
	author = {Clark, Herbert H.},
	month = may,
	year = {1996},
}

@misc{shen_position_2025,
    title = {Position: {Towards} {Bidirectional} {Human}-{AI} {Alignment}},
    shorttitle = {Position},
    url = {http://arxiv.org/abs/2406.09264},
    doi = {10.48550/arXiv.2406.09264},
    urldate = {2026-01-13},
    publisher = {arXiv},
    author = {Shen, Hua and Knearem, Tiffany and Ghosh, Reshmi and Alkiek, Kenan and Krishna, Kundan and Liu, Yachuan and Ma, Ziqiao and Petridis, Savvas and Peng, Yi-Hao and Qiwei, Li and Rakshit, Sushrita and Si, Chenglei and Xie, Yutong and Bigham, Jeffrey P. and Bentley, Frank and Chai, Joyce and Lipton, Zachary and Mei, Qiaozhu and Mihalcea, Rada and Terry, Michael and Yang, Diyi and Morris, Meredith Ringel and Resnick, Paul and Jurgens, David},
    month = sep,
    year = {2025},
    note = {arXiv:2406.09264 [cs]},
}

\end{document}